\pdfoutput=1

\documentclass[11pt]{article}
\usepackage{booktabs}
\usepackage{float}

\usepackage{acl}
\usepackage{tcolorbox}

\usepackage{times}
\usepackage{latexsym}

\usepackage[T1]{fontenc}

\usepackage[utf8]{inputenc}

\usepackage{microtype}
\usepackage{inconsolata}

\usepackage{booktabs} 
\usepackage{multirow} 
\usepackage{amsmath}  
\usepackage{graphicx} 
\usepackage{enumitem}
\usepackage{tabularx}
\title{HABIB\_TAZ at SemEval-2026 Task 11: Disentangling Formal Logic from Content via Synthetic Training and Multi-Objective Optimization}

\author{
  Abdullah Shaikh\thanks{\hspace{0.15cm}Equal contribution.},~ 
  Zain Naqi\footnotemark[1],~ 
  Taha Zahid\footnotemark[1],~ 
  Sandesh Kumar\footnotemark[1],~ 
  Abdul Samad \\
  Dhanani School of Science \& Engineering \\
  Habib University, Pakistan \\
  \texttt{\{as09245, zn09224, tz09220\}@st.habib.edu.pk} \\
  \texttt{\{sandesh.kumar, abdul.samad\}@sse.habib.edu.pk}
}
\begin{document}
\maketitle
\thispagestyle{empty}

\begin{abstract}
While Large Language Models (LLMs) excel in many general NLP tasks, their formal reasoning capabilities are often compromised by content effects, demonstrating a measurable bias towards real-world plausibility. In this paper, we present our system for SemEval-2026 Task 11, which evaluates the ability of models to disentangle formal logic from content across 12 languages with and without distractor premises. We address this challenge using mDeBERTa-v3 networks fine-tuned on a synthetic, rule-based dataset of syllogistic schemes to avoid the semantic noise of LLM-augmented data. To explicitly decouple plausibility from logical structure, our training pipeline employs a multi-objective loss function combining Adaptive Group Distributionally Robust Optimization (DRO), a scheduled differentiable bias penalty, and KL-Divergence consistency regularization. Our system achieved \#1 ranks and perfect Ranking Scores (100.0) with 0.00\% bias and 100.0\% accuracy on Subtask 1 (English), Subtask 2 (Noisy English), and Subtask 3 (Multilingual). On the highly complex Subtask 4 (Noisy Multilingual), the system achieved the 6th rank with 89.06\% Accuracy and F1-score, alongside a limited 2.89\% Bias and a 37.78 Ranking Score. Our dataset generation engine and codebase are publicly available to facilitate future work on robust logical reasoning.
\end{abstract}

\section{Introduction}
While Large Language Models (LLMs) \cite{Cheng2025, Wan2025, Quan2024} excel in general NLP, they struggle with "content effects", overestimating the logical validity of arguments aligned with world knowledge \cite{dasgupta2024languagemodelshumanlikecontent, Valentino2025, Kim2024}. This entanglement hinders the development of trustworthy AI that prioritizes logical robustness \cite{huang2024trustllmtrustworthinesslargelanguage}. We address this in SemEval-2026 Task 11 \cite{valentino-etal-2026-semeval}, aiming to assess the formal validity of syllogistic arguments across 12 diverse languages\footnote{English (en), German (de), Spanish (es), French (fr), Italian (it), Dutch (nl), Portuguese (pt), Russian (ru), Chinese (zh), Swahili (sw), Bengali (bn), and Telugu (te).}, independent of their plausibility (alignment with world knowledge). Furthermore, the task requires models to exhibit robustness by identifying relevant premises amidst irrelevant noise.

Our system utilizes mDeBERTa-v3 backbones \cite{he2021debertav3} pre-trained on NLI tasks \cite{laurer2022less}. To ground the architecture in formal logic, we curate a synthetic dataset via a rule-based engine following syllogistic schemes \cite{Bertolazzi2024}, avoiding the noise of LLM-augmented data. We optimize via a multi-objective function integrating Group DRO \cite{sagawa2020distributionallyrobustneuralnetworks}, a scheduled bias-weighting parameter $\lambda$, and KL Divergence to decouple formal reasoning from plausibility, and to enforce structural invariance across complex linguistic variations.

Our code and data generation scripts are publicly available to facilitate further research.\footnote{\url{https://github.com/TahaZahid05/SemEval26-Task-11}}

\section{Task Description}
The task requires classifying the logical validity of a conclusion ($C$) given premises ($P$) across four subtasks. Evaluation targets the 2$\times$2 intersection of \textbf{Validity} and \textbf{Plausibility} (Table \ref{tab:examples}).

\begin{table*}[t]
\centering
\begin{tabularx}{\textwidth}{@{}lXcc@{}}
\toprule
\textbf{Quadrant} & \textbf{Example (Premises $\rightarrow$ Conclusion)} & \textbf{Validity} & \textbf{Plausibility} \\ \midrule
Valid Plausible & All cats are animals. Luna is a cat $\rightarrow$ \textbf{Luna is an animal.} & 1 & 1 \\
Valid Implausible & All birds can swim. Eagles are birds $\rightarrow$ \textbf{Eagles can swim.} & 1 & 0 \\
Invalid Plausible & All dogs have fur. Some pets have fur $\rightarrow$ \textbf{Some pets are dogs.} & 0 & 1 \\
Invalid Implausible & All rocks are soft. All clouds are soft $\rightarrow$ \textbf{Rocks are clouds.} & 0 & 0 \\ \bottomrule
\end{tabularx}
\caption{Task structure and logical quadrants. In "Noisy" tracks (ST2/ST4), models must also identify the subset of relevant premises, filtering out distractor sentences like "The sky is blue."}
\label{tab:examples}
\end{table*}

We participated in all four subtasks: \textbf{ST1} (English), \textbf{ST2} (English with irrelevant premises), \textbf{ST3} (12-language multilingual), and \textbf{ST4} (12-language multilingual with irrelevant premises). ST2 and ST4 require joint validity classification and relevant premise selection. The primary metric is the \textbf{Ranking Score} ($S_{rank}$), which divides a performance component $P$ by a logarithmic penalty on the \textbf{Total Content Effect (TCE)} \cite{valentino-etal-2026-semeval}. $P$ is Accuracy for ST1/ST3 and the average of Accuracy and F1 for ST2/ST4. TCE measures bias by comparing accuracy across the validity--plausibility quadrants; lower TCE indicates higher logical robustness.

The training dataset provided contained 960 English data points balanced across quadrants (VP: 25.0\%, VI: 25.0\%, IP: 24.4\%, II: 25.6\%).

\section{Related Work}
Recent work on logical reasoning in LLMs can be categorized along four directions, following the taxonomy of \cite{Cheng2025}.

\paragraph{Neuro-Symbolic Approaches}
\cite{Quan2024} integrates external theorem provers to verify LLM-generated natural language explanations. By iteratively formalizing explanations and validating them with a logic solver, their feedback loop ensures logical soundness and completeness.

\paragraph{Prompt-Based Methods}
\cite{Wan2025} propose a Syllogistic-Reasoning Framework of Thought (SR-FoT), extending Chain-of-Thought prompting to a multi-stage syllogistic reasoning process. By generating major and minor premises before the conclusion, SR-FoT enforces strict information control, reducing errors and improving logical consistency.

\paragraph{Model Optimization}
\cite{Valentino2025} proposes Activation Steering to disentangle content plausibility from logical validity within LLM hidden representations. Using probing techniques to identify content-sensitive layers, they apply conditional steering to guide the model toward formal reasoning. Similarly, \cite{Wu2024} proposes a Logical Control Framework that separates content and logic subspaces via contrastive learning, nudging representations toward logical validity during inference without altering content.

\paragraph{Analytical Studies}
\cite{Kim2024} identifies distinct attention circuits responsible for formal reasoning versus content-dependent processing, revealing that certain attention heads are both necessary and sufficient for logical inference. On the evaluation side, \cite{Patel2024} introduces Multi-LogiEval, demonstrating that LLM accuracy degrades as the number of premises and reasoning steps increases.


\section{System Overview}
Our system is designed to decouple semantic patterns from logical structure through synthetic data generation, bias-aware optimization, and leveraging the structural attention of mDeBERTa-v3 backbones.

\subsection{Architecture}
We utilize mDeBERTa-v3 \cite{he2021debertav3} as our backbone, initialized from NLI-pretrained checkpoints \cite{laurer2022less}: \texttt{mDeBERTa-\allowbreak v3-\allowbreak base-\allowbreak xnli-\allowbreak multilingual-\allowbreak nli-\allowbreak 2mil7} for ST1 and ST3, \texttt{DeBERTa-\allowbreak v3-\allowbreak large-\allowbreak mnli-\allowbreak fever-\allowbreak anli-\allowbreak ling-\allowbreak wanli} for ST2, and \texttt{microsoft/\allowbreak mdeberta-\allowbreak v3-\allowbreak base} for ST4.\footnote{All checkpoints are available on HuggingFace under the \texttt{MoritzLaurer} namespace, except for ST4 which uses a Microsoft checkpoint.} The larger backbone for ST2 proved necessary for disentangling noisy premises. For ST1/ST3, a classification head consumes the \texttt{[CLS]} token for binary validity classification. For ST2/ST4, we use two specialized heads:
\begin{itemize}
    \item \textbf{Classification Head:} Predicts logical validity from the \texttt{[CLS]} embedding.
    \item \textbf{Premise Selection Head:} Uses span-based pooling between \texttt{[SEP]} tokens to extract premise representations. We apply Masked Mean Pooling to each span, passing the resulting fixed-size vectors to a sigmoid layer for relevance probability.
\end{itemize}

\subsection{Data Strategy}
To overcome the scarcity of the provided dataset ($N=960$), we developed a rule-based engine to generate synthetic data based on the 256 Aristotelian moods and figures \cite{Bertolazzi2024}. 

\paragraph{Vocabulary and Plausibility} 
We select nouns from WordNet \cite{miller1995wordnet} hypernym/hyponym chains using the NLTK library \cite{bird2009natural}. Plausible samples preserve hierarchical ordering (e.g., $car \subset vehicle \subset artifact$) for nouns, while implausible samples use cross-chain or random nouns. To further reduce content reliance, we introduce \textbf{Symbolic Data} (using gibberish/symbolic variables) and \textbf{Complex Rephrasing} (e.g., mapping "All $x$ are $y$" to "Every $x$ is, without exception, a $y$.").

\paragraph{Relevant Premises Labeling}
For invalid syllogisms in ST2 and ST4, we retain the indices of premises that provide partial entailment. This prevents the model from treating premise relevance as a purely binary consequence of overall validity, encouraging structural over-fitting to logical forms.

\paragraph{Multilingual Expansion}
We first generated English samples, then translated nouns via DeepL\footnote{\url{https://www.deepl.com/}}, and then translated the structure using language-specific rule-based templates. This ensures that the quantifiers (e.g., "No", "Some") are not modified across all 12 target languages.

\paragraph{Syntactic Robustness (HANS)}
To ensure the models do not exploit surface-level syntactic heuristics (e.g., lexical overlap) during inference, we additionally integrate samples from the Heuristic Analysis for NLI Systems (HANS) dataset \cite{mccoy2019rightwrongreasonsdiagnosing} into our robust training distributions.

\subsection{Bias-Aware Optimization}
To optimize for the Ranking Score, we use a multi-objective function:
\begin{equation}
    \mathcal{L}_{total} = \mathcal{L}_{DRO} + \lambda_{bias} \mathcal{L}_{bias} + \gamma \mathcal{L}_{KL}
\end{equation}

\paragraph{Adaptive Group DRO and Scheduling}
Instead of Cross Entropy Loss, which prioritizes global performance, we utilize \textbf{Group Distributionally Robust Optimization (DRO)} \cite{sagawa2020distributionallyrobustneuralnetworks} to ensure that the model stays consistent by focusing more on low-performing subgroups. We define $G=6$ subgroups: $\{VP, VI, IP, II\}$ and two \textit{Symbolic} groups. For multilingual tasks (ST3/ST4), each of the six subgroups is further divided into 12 subgroups, one for each language, resulting in a total of 72 subgroups. The model optimizes $\mathcal{L}_{DRO} = \sum_{g=1}^{G} q_g \mathcal{L}_g$, where group weights $q_g$ are updated adaptively:
\begin{equation}
    q_g^{(t+1)} \propto q_g^{(t)} \exp(\eta \mathcal{L}_g^{(t)})
\end{equation}
 To ensure the model establishes a logical baseline before unbiasing, we implement \textbf{Dynamic Scheduling}: $\lambda_{bias}$ and the DRO step-size $\eta$ are initialized at 0 for the first epoch and incrementally increased. This prevents the optimization from collapsing into sub-optimal local minima where the model might ignore semantic content before mastering formal syllogistic structures.

\paragraph{Differentiable Bias Penalty}
To align the training objective with the Content Effect (CE) evaluation metrics, we formulate a loss component using sigmoid confidence scores $\hat{y}$. Let $\hat{y}_{v,pl}$ be the model's confidence scores for samples with validity $v \in \{0, 1\}$ and plausibility $pl \in \{0, 1\}$. We calculate:
\begin{align*}
    \bar{\hat{y}}_{Intra} &= \frac{1}{2} \sum_{p \in \{0, 1\}} | \text{mean}(\hat{y}_{1,p}) - \text{mean}(\hat{y}_{0,p}) | \\
    \bar{\hat{y}}_{Cross} &= \frac{1}{2} \sum_{v \in \{0, 1\}} | \text{mean}(\hat{y}_{v,1}) - \text{mean}(\hat{y}_{v,0}) |
\end{align*}
The final $\mathcal{L}_{bias}$ is the mean of these effects. This approach allows the gradient to penalize uneven confidence distributions across quadrants, effectively "forcing" the model to maintain high logical certainty regardless of real-world plausibility.

\paragraph{Consistency via KL-Divergence}
To ensure robustness to linguistic variation, we minimize the KL-Divergence ($\mathcal{L}_{KL}$) \cite{kullback1951information} between the output distributions of a simple syllogism (teacher) and its complex rephrasing (student). For example, mapping the simple form "All $A$ are $B$" to the complex "If something is $A$, then it is $B$". This $\min \text{KL}(P_{simple} || P_{complex})$ optimization forces the model to align the semantic representation of complex conditionals with their fundamental logical forms, thereby preventing reliance on lexical shortcuts. This approach deliberately avoids converting complex syllogisms into a normalized abstract form, as such conversion would either require LLM-based paraphrasing, introducing probabilistic errors, or a deterministic rule-based normalizer that itself constitutes a non-trivial engineering overhead.

\subsection{Data Splits and Preprocessing}

We use a stratified 80/10/10 split (Train/Validation/Test) for both the training dataset provided by the curators \cite{valentino-etal-2026-semeval}, and our augmented dataset. Stratification is performed across to ensure balanced logical and semantic representation in every split.

\subsection{Dataset Composition}

Our augmentation pipeline expanded the fully balanced initial dataset ($N=960$) to include a more complete set of logical forms and variations: $23,524$ total samples for ST1, $27,988$ for ST2, 388,992 for ST3, and $648,632$ for ST4. All subsets remain balanced across the four logic/plausibility quadrants and also across languages for multilingual subtasks.

\subsection{Hyperparameters}

We use the AdamW optimizer \cite{loshchilov2017decoupled} with a linear schedule and a warmup ratio of 0.06. To account for the stability of the backbone, we apply a differential learning rate ($\text{LR}_{backbone} = 0.1 \times \text{LR}_{head}$). Specific subtask hyperparameter configurations are detailed in Appendix \ref{sec:appendix_hyperparams}.

\subsection{Implementation and Hardware}

The system is implemented in PyTorch \cite{paszke2019pytorch} using HuggingFace Transformers \cite{wolf-etal-2020-transformers}. Training used NVIDIA Tesla P100 (16GB) and T4 Tensor Core (16GB) GPUs with Mixed Precision (FP16) and a Gradient Scaler.


\begin{table*}[h!]
\centering
\small
\begin{tabular}{llcccc}
\toprule
\textbf{Subtask} & \textbf{Configuration} & \textbf{Acc ($\uparrow$)} & \textbf{F1-P ($\uparrow$)} & \textbf{TCE ($\downarrow$)} & \textbf{$S_{rank}$ ($\uparrow$)} \\ \midrule
\multirow{6}{*}{\textbf{ST1}}
& Vanilla (Baseline) & 100.0 & -- & 0.0 & 100.0 \\
& Bias & 100.0 & -- & 0.0 & 100.0 \\
& DRO + Bias & 100.0 & -- & 0.0 & 100.0 \\
& MSE + DRO + Bias$^\dagger$ & 100.0 & -- & 0.0 & 100.0 \\
& FreeLB + MSE + DRO + Bias$^\dagger$ & 100.0 & -- & 0.0 & 100.0 \\
& \textbf{KL + DRO + Bias (Final)} & \textbf{100.0} & -- & \textbf{0.0} & \textbf{100.0} \\ \midrule

\multirow{8}{*}{\textbf{ST2}}
& Vanilla (Baseline) & 98.95 & 97.89 & 2.13 & 45.99 \\
& DRO + Bias & 97.89 & 95.79 & 4.26 & 36.42 \\
& Bias & 98.42 & 96.84 & 3.19 & 40.13 \\
& MSE + Bias$^\dagger$ & 98.42 & 96.84 & 3.19 & 40.13 \\
& MSE + DRO + Bias$^\dagger$ & 97.89 & 96.84 & 2.20 & 45.03 \\
& FreeLB + MSE + DRO + Bias$^\dagger$ & 99.47 & 98.95 & 1.06 & 57.53 \\
& KL + DRO + Bias & 99.47 & 98.95 & 1.06 & 57.53 \\
& \textbf{KL + DRO + Bias (Large)} & \textbf{100.0} & \textbf{100.0} & \textbf{0.0} & \textbf{100.0} \\ \midrule

\multirow{8}{*}{\textbf{ST3}}
& Vanilla (Baseline) 
& 91.15 & -- & 8.29 & 28.23 \\

& Bias + Hans + DRO + KL (non symbolic) 
& 98.44 & -- & 2.08 & 46.30 \\

& Bias + Hans + DRO + MSE$^\dagger$ (non symbolic) 
& 98.44 & -- & 2.08 & 46.30 \\

& Bias (non-symbolic) 
& 99.48 & -- & 1.09 & 57.31 \\

& Bias 
& 98.96 & -- & 1.04 & 57.74 \\

& Bias + Hans + DRO + KL
& 99.48 & -- & 1.04 & 58.05 \\

& Bias + Hans + DRO + MSE$^\dagger$ 
& 99.48 & -- & 1.04 & 58.05 \\

& Bias + HANS + DRO
& 100.00 & -- & 0.00 & 100.00 \\ 

& \textbf{Bias + HANS} 
& \textbf{100.00} & -- & \textbf{0.00} & \textbf{100.00} \\ 

\midrule

\multirow{8}{*}{\textbf{ST4}}
& Vanilla (Baseline, MT) & 85.93 & 81.77 & 7.22 & 26.98 \\
& DRO + Bias (MT) & 84.89 & 79.68 & 8.26 & 25.50 \\
& Bias (MT) & 84.37 & 80.20 & 8.24 & 25.98 \\
& KL + DRO + Bias (MT) & 88.02 & 85.93 & 4.99 & 31.16 \\ 
& KL + DRO + Bias (ET) & 89.06 & 91.66 & 5.03 & 32.29 \\ 
& ST2 Best Model (ET) & 89.06 & 89.06 & 2.89 & 37.78 \\ 
& DRO + Bias (ET) & 89.06 & 88.02 & 2.30 & 40.35 \\
& \textbf{Bias (ET)} & \textbf{89.06} & \textbf{90.62} & \textbf{2.21} & \textbf{41.43} \\
\bottomrule
\end{tabular}
\caption{System performance across all experimented subtasks and configurations. $\dagger$ indicates experimental loss components. F1-P denotes the F1 score for premise selection. $S_{rank}$ for ST2-4 is the combined ranking score. For ST4, (MT) represents the original multilingual test data given by SemEval Task curators, while (ET) represents its English translation by ChatGPT \cite{openai2026gpt52}.}
\label{tab:ablation_results}
\end{table*}

\section{Results and Analysis}

\subsection{Main Quantitative Findings}
Table \ref{tab:ablation_results} provides the complete set of experimental configurations and ablation studies conducted across all four subtasks, with the best performing models per subtask highlighted. Our primary objective was to maximize the Ranking Score ($S_{rank}$) by minimizing the Total Content Effect (TCE) and maximizing the performance component $P$. We additionally explored a couple of other experimental loss components, including \textbf{Mean Squared Error (MSE)} and \textbf{FreeLB} \cite{DBLP:journals/corr/abs-1909-11764}; while these provided competitive stability, they were not included in the final system overview as they were not in our best results. \textbf{Details on these experimental components can be found in Appendix \ref{sec:appendix_loss}}.

As shown in the ST1 results, all configurations achieved perfect performance, suggesting that for ST1, the mDeBERTa-v3 backbone combined with our augmented dataset is sufficient. 

However, the complexity of ST2 revealed performance gaps. Our final configuration using \textbf{KL + DRO + Bias} on the large model backbone achieved a perfect score of 100.0, successfully disentangling formal reasoning from noisy premises and content bias.

For ST3, using the bias alone produced near-perfect accuracy. Both \textbf{Bias + HANS} and \textbf{Bias + HANS + DRO} achieved a perfect score of 100.00. The addition of KL or MSE regularization failed to preserve this performance, suggesting that bias loss combined with HANS (with or without DRO) is sufficient to disentangle reasoning from content bias.

For the highly complex ST4, the \textbf{Bias} configuration achieved a Ranking Score of 37.78 during official evaluation. Notably, post-competition analysis revealed a drastic performance gap when evaluating on the original multilingual test set compared to its English translation generated via ChatGPT \cite{openai2026gpt52}. As shown in Table \ref{tab:ablation_results}, the \textbf{DRO + Bias} model scored 25.50 on multilingual data but jumped to 40.35 on translated English data. A similar $\approx 60\%$ rank increase occurs for the standalone \textbf{Bias} model. Interestingly, as we discuss further in Section 5.2, applying KL-Divergence consistency regularization significantly closes this cross-lingual gap.

\subsection{Ablation and Error Analysis}
For ST2, our experiments indicate that while the Vanilla model was able to achieve some success, it is susceptible to content effects ($2.13\%$ TCE). 

Surprisingly, adding only Differentiable Bias (40.13) or DRO + Differential Bias (36.42) degraded the performance compared to vanilla. DRO + Differential Bias showed the highest content effect (4.26\%), which suggests that aggressive optimization can dominate the loss function and degrade performance. This finding underscores the importance of carefully designed multi-component systems rather than isolated regularization techniques. We found that an additional component, such as KL-Divergence or MSE, acted as a regularization anchor ensuring that even as the model is pushed to be unbiased, it remains grounded in the underlying formal syllogistic structure.

Our complete system (KL + DRO + Bias) achieved a 57.53 ranking score on the base model, representing a 25\% improvement over vanilla. The FreeLB + MSE + DRO + Bias configuration achieved identical performance, indicating that both KL divergence and FreeLB provide similar consistency regularization benefits. However, FreeLB is computationally more expensive since it runs multiple forward passes per input.

Scaling the architecture from the Base to the Large model in the ST2 (KL + DRO + Bias) configuration yielded a perfect ranking score. This suggests that the capacity to abstract logical relationships from noisy premises scales directly with model parameters. Furthermore, while MSE and FreeLB provided stability during baseline experimentation, KL-Divergence consistency regularization proved more efficient and equally effective when generalizing to the Large backbone.

For ST3, the Vanilla model achieved a strong accuracy of 91.15\%, but remained highly susceptible to content effect (8.29\% TCE).

Introducing only the Differentiable Bias significantly improved the performance, accuracy increased to 98.96, while content effect dropped sharply to 1.04\%, which suggests the effectiveness of bias-aware loss in mitigating the content effect. However, the most substantial jump occurred with the inclusion of the HANS dataset. The \textbf{Bias + HANS} configuration achieved a perfect 100.00 final score. Integrating DRO into this configuration maintained the perfect performance, indicating the model's stability across different data distributions.

Interestingly, unlike the trends observed in ST2, additional consistency regularizers, KL-Divergence and MSE, failed to sustain the perfect score. Instead, they degraded the ranking scores and reintroduced the content effect. Both KL and MSE yielded identical results, suggesting that the model had already reached the optimal state and additional regularization may have introduced unnecessary constraints. 

Furthermore, a critical takeaway from the ST3 results is the comparison between models trained without symbolic data and those trained with symbolic data. The models trained with symbolic data consistently outperformed the ones trained without it, suggesting that symbolic datasets contributed heavily to generalization. By training on abstract symbols rather than just real-world nouns, the model effectively disentangles logical reasoning from content bias. 

For ST4, a pattern similar to ST2 emerges: DRO + Bias and Bias alone perform worse than Vanilla on multilingual test data (MT). To diagnose this, we translated the multilingual test set to English (ET) via GPT-5.2 \cite{openai2026gpt52}. The results reveal a significant cross-lingual gap: DRO + Bias scored 25.50 on MT but 40.35 on ET, and Bias alone scored 25.98 vs.\ 41.43, a $\approx 60\%$ rank increase, indicating that the core failure stems from (i) difficulty generalizing multilingual premise structures and/or (ii) ``translationese'' noise in mDeBERTa-v3's pre-training data \cite{artetxe-etal-2020-translation}. In contrast, the \textbf{KL + DRO + Bias} configuration achieved near-identical scores on both test sets ($S_{rank}$ of 31.16 on MT vs.\ 32.29 on ET), showing that KL-Divergence consistency regularization makes the model robust to cross-lingual variation by anchoring on simple English syllogisms and minimizing output divergence across all 12 language variants. However, this requires 24 forward passes per data point; due to compute constraints, training data was reduced to $\approx 4\%$, suggesting further gains are achievable with more data.

\section{Conclusion}
We presented a system for disentangling formal logical reasoning from content effects in syllogistic argument classification. Our approach combines a rule-based synthetic data generation engine covering 256 Aristotelian syllogistic schemes across 12 languages with a multi-objective loss function integrating a scheduled differentiable bias penalty, Adaptive Group DRO, and KL-Divergence consistency regularization. The system achieved perfect Ranking Scores (100.0) with 0.00\% bias on ST1, ST2, and ST3, confirming that bias-aware optimization with synthetic data can effectively isolate formal reasoning from real-world plausibility. On the most challenging subtask ST4, the system ranked 6th with 89.06\% accuracy and 2.89\% bias. The primary direction for future work is training the full \textbf{KL + DRO + Bias} pipeline on the complete ST4 dataset (648,632 samples), which was reduced to $\approx$4\% during the competition due to compute constraints, and which demonstrated the strongest cross-lingual robustness among all configurations.

\bibliography{custom}

\appendix
\section{Experimental Loss Components}
\label{sec:appendix_loss}
As noted in Section 5.1, while our final winning configuration relied on KL-Divergence for consistency regularization and avoided adversarial perturbations to maintain computational efficiency on the Large backbone, our detailed ablation study (Table \ref{tab:ablation_results}) reports metrics on two additional experimental loss components: Mean Squared Error (MSE) consistency and FreeLB (Free Large-Batch Adversarial Training).
\subsection{Mean Squared Error (MSE) Consistency Regularization}
Before adopting KL-Divergence, we experimented with Mean Squared Error (MSE) to enforce structural invariance. Under this configuration, every data point was structured to contain both a \texttt{syllogism\_simple} and a \texttt{syllogism\_complex} text variant. For the multilingual subtasks (ST3/ST4), this was expanded further such that a single data point could contain up to 24 language-specific simple and complex variations of the same underlying logical structure. 
During the forward pass, the model generated logits for both the simple constraint ($z_{simple}$) and the complex constraint ($z_{complex}$). The MSE consistency penalty was formulated as:
\begin{equation}
    \mathcal{L}_{MSE} = \frac{1}{N} \sum_{i=1}^{N} (z_{simple}^{(i)} - z_{complex}^{(i)})^2
\end{equation}
This loss component successfully anchored the model, ensuring that the confidence scores for semantically complex variations closely mapped to their fundamental, simple logical counterparts. While MSE provided competitive stability (Ranking Score 45.03 in ST2), we ultimately found that minimizing the KL-Divergence between the output probability distributions was more effective at preventing the model from relying on lexical shortcuts than minimizing the absolute difference between the raw logits. 
\subsection{Free Large-Batch Adversarial Training (FreeLB)}
To further test the robustness of the model against grammatical noise and irrelevant premises, we implemented FreeLB \cite{DBLP:journals/corr/abs-1909-11764}. FreeLB adds adversarial perturbations to word embeddings during training, forcing the model to make consistent predictions even when the input representations are subjected to small, continuous gradient-based attacks. 
We applied the standard FreeLB algorithm, executing multiple forward passes per input batch. In each inner step, we perturbed the embeddings by a constrained magnitude $\epsilon$ and calculated the loss, updating the model parameters based on the accumulated gradients of the adversarially perturbed inputs. 
As shown in Table \ref{tab:ablation_results}, the combination of FreeLB + MSE + DRO + Bias achieved a Ranking Score of 57.53 on the Base model, which was identical to the performance of our KL + DRO + Bias configuration. FreeLB successfully stabilized the aggressively debiased model. However, because FreeLB requires multiple forward passes per iteration, it is computationally expensive. When transitioning our architecture to the Large mDeBERTa-v3 backbone to maximize the Ranking Score, we dropped FreeLB in favor of the computationally lighter KL-Divergence approach, which yielded the same consistency benefits without multiplying the training time.
\section{Extended Data Generation Procedure}
\label{sec:appendix_data}

As described in Section 3.2, our synthetic dataset was generated programmatically. To isolate the effects of structural noise, linguistic complexity, and consistency pairing, the data was generated via independent script executions rather than a monolithic pipeline. 

\subsection{Base Generation}
The baseline dataset was generated using a 256-rule Aristotelian logic engine. This initial procedure outputs 'Simple' English syllogisms encompassing non-symbolic nouns, symbolic variables, and both 3-variable and 4-variable logical constraints. All subsequent augmentation scripts act upon this foundational subset.

\subsection{Irrelevant Premise Injection}
For robustness configurations (e.g., ST2/ST4), modifier scripts were executed over the base English data to inject structural noise:
\begin{enumerate}
    \item \textbf{Premise Removal:} The modifier script discarded a structurally critical premise from a designated subset of valid chains, directly invalidating the logic. 
    \item \textbf{Distractor Generation:} The sequence then generated logically disjoint distractor premises (e.g., inserting a premise regarding \textit{"clouds"} into a syllogism concerning \textit{"dogs"}) and concatenated them to the input text.
\end{enumerate}

\subsection{Paired Semantic Augmentation}
Consistency optimization (e.g., KL-Divergence or MSE) requires identically paired "simple" and "complex" variants of underlying logical forms. For English datasets, a secondary complex-rephrasing script was applied to the base output, applying synonym replacement and structural permutations while maintaining parity with the corresponding simple input matrices.

\subsection{Multilingual Translation}
For the 11 non-English language tracks (ST3/ST4), translation followed two independent pathways depending on the required complexity constraint:
\begin{enumerate}
    \item \textbf{Simple Translation:} The complete English input text (with or without injected noise) was translated into the target languages via the DeepL API. 
    \item \textbf{Complex Translation:} To execute complex rephrasing natively across languages, a specialized script isolated and translated only the semantic nouns from the English text. These localized noun-sets were subsequently passed to 11 language-specific generator scripts (e.g., \texttt{ComplexSyllogisms\_French.ipynb}), which applied complex phrasing mappings independently. Because symbolic datasets lack semantic noun dependencies, they bypassed this translation step entirely.
\end{enumerate}

\section{Hyperparameter Configurations}
\label{sec:appendix_hyperparams}

\subsection{Subtask 1 (ST1) Configurations}
\begin{table}[H]
\centering
\small
\begin{tabular}{@{}ll@{}}
\toprule
\textbf{Hyperparameter} & \textbf{Value} \\ \midrule
Model & mDeBERTa-v3-base \\
Batch Size & 32 \\
Gradient Accumulation Steps & 1 \\
Epochs & 10 \\
Learning Rate & $2 \times 10^{-5}$ \\
Warmup Ratio & 0.06 \\
Weight Decay & 0.01 \\
Max Sequence Length & 96 \\
Dropout Rate & 0.1 \\
Max $\lambda_{\text{bias}}$ & 2.0 \\
$\gamma$ (KL Consistency) & 0.5 \\
DRO Warmup & 1 Epoch \\
Early Stopping Patience & 2 \\
Early Stopping Min $\Delta$ & 0.001 \\
Mixed Precision (FP16) & True \\ \bottomrule
\end{tabular}
\caption{Hyperparameters for Subtask 1 (ST1).}
\end{table}

\subsection{Subtask 2 (ST2) Configurations}
\begin{table}[H]
\centering
\small
\begin{tabular}{@{}ll@{}}
\toprule
\textbf{Hyperparameter} & \textbf{Value} \\ \midrule
Model & DeBERTa-v3-large \\
Batch Size & 6 \\
Gradient Accumulation Steps & 10 \\
Effective Batch Size & 60 \\
Epochs & 10 \\
Learning Rate & $5 \times 10^{-6}$ \\
Warmup Ratio & 0.06 \\
Weight Decay & 0.01 \\
Max Sequence Length & 232 \\
Premise Buffer Size & 9 \\
Dropout Rate & 0.1 \\
Max $\lambda_{\text{bias}}$ & 2.0 \\
$\gamma$ (KL Consistency) & 0.5 \\
DRO Warmup & 1 Epoch \\
Early Stopping Patience & 2 \\
Early Stopping Min $\Delta$ & 0.001 \\
Mixed Precision (FP16) & True \\ \bottomrule
\end{tabular}
\caption{Hyperparameters for Subtask 2 (ST2).}
\end{table}

\subsection{Subtask 3 (ST3) Configurations}
\begin{table}[H]
\centering
\small
\begin{tabular}{@{}ll@{}}
\toprule
\textbf{Hyperparameter} & \textbf{Value} \\ \midrule
Model & mDeBERTa-v3-base-xnli \\
Batch Size & 144 \\
Gradient Accumulation Steps & 1 \\
Effective Batch Size & 144 \\
Epochs & 3 \\
Learning Rate & $2 \times 10^{-5}$ \\
Warmup Ratio & 0.06 \\
Weight Decay & 0.01 \\
Max Sequence Length & 96 \\
Dropout Rate & 0.1 \\
Max $\lambda_{\text{bias}}$ & 2.5 \\
DRO Warmup & 1 Epoch \\
Early Stopping Patience & 2 \\
Early Stopping Min $\Delta$ & 0.001 \\
Mixed Precision (FP16) & True \\
Pooling type & cls \\\bottomrule
\end{tabular}
\caption{Hyperparameters for Subtask 3 (ST3).}
\end{table}

\subsection{Subtask 4 (ST4) Configurations}
\begin{table}[H]
\centering
\small
\begin{tabular}{@{}ll@{}}
\toprule
\textbf{Hyperparameter} & \textbf{Value} \\ \midrule
Model & microsoft/mdeberta-v3-base \\
Batch Size & 72 \\
Gradient Accumulation Steps & 1 \\
Effective Batch Size & 72 \\
Epochs & 3 \\
Learning Rate & $5 \times 10^{-6}$ \\
Warmup Ratio & 0.06 \\
Weight Decay & 0.01 \\
Max Sequence Length & 265 \\
Premise Buffer Size & 16 \\
Dropout Rate & 0.1 \\
Max $\lambda_{\text{bias}}$ & 2.0 \\
DRO Warmup & 1 Epoch \\
Early Stopping Patience & 2 \\
Early Stopping Min $\Delta$ & 0.001 \\
Mixed Precision (FP16) & True \\ \bottomrule
\end{tabular}
\caption{Hyperparameters for Subtask 4 (ST4).}
\end{table}

\end{document}